\documentclass[twoside]{article}
\usepackage[accepted]{arxiv}

\usepackage[numbers,sort&compress]{natbib}

\usepackage[pdftex]{graphicx}
\usepackage{amsmath}
\usepackage{amssymb}

\usepackage{subcaption}
\usepackage{multirow}
\usepackage[abbreviations]{foreign}

\usepackage[utf8]{inputenc} 
\usepackage[T1]{fontenc}    
\usepackage{hyperref}       
\usepackage{url}            
\usepackage{booktabs}       
\usepackage{amsfonts}       
\usepackage{nicefrac}       
\usepackage{microtype}      
\usepackage{algorithm}
\usepackage{algpseudocode}
\usepackage{cuted}

\makeatletter
\renewcommand{\ALG@beginalgorithmic}{\small}
\makeatother

\DeclareMathOperator*{\E}{\mathbb{E}}

\begin{document}

\twocolumn[

\aistatstitle{Probabilistic Adaptive Computation Time}

\aistatsauthor{Michael Figurnov$^{1}$ \qquad Artem Sobolev$^{*2}$  \qquad Dmitry Vetrov$^{1,3}$}
\runningauthor{Michael Figurnov, Artem Sobolev, Dmitry Vetrov}
\aistatsaddress{$^1$National Research University Higher School of Economics \qquad $^{2}$ Luka Inc. \qquad $^3$Yandex \\
\texttt{\small michael@figurnov.ru \qquad stdio@artem.sobolev.name \qquad vetrovd@yandex.ru }
}
]

\begin{abstract}

We present a probabilistic model with discrete latent variables that control the computation time in deep learning models such as ResNets and LSTMs.
A prior on the latent variables expresses the preference for faster computation.
The amount of computation for an input is determined via amortized maximum a posteriori (MAP) inference.
MAP inference is performed using a novel stochastic variational optimization method.
The recently proposed Adaptive Computation Time mechanism can be seen as an ad-hoc relaxation of this model.
We demonstrate training using the general-purpose Concrete relaxation of discrete variables.
Evaluation on ResNet shows that our method matches the speed-accuracy trade-off of Adaptive Computation Time, while allowing for evaluation with a simple deterministic procedure that has a lower memory footprint.

\end{abstract}

\renewcommand*{\thefootnote}{\fnsymbol{footnote}}
\footnotetext[1]{This work was done while studying at NRU HSE.}
\renewcommand*{\thefootnote}{\arabic{footnote}}

\section{Introduction}

In the past years, deep learning models have become significantly deeper and more computationally expensive.
As evident from the ImageNet competition results \cite{krizhevsky2012imagenet,simonyan2014verydeep,szegedy2015inception,he2016deep}, increasing the depth of computer vision models indeed leads to improved results.
However, such expensive models are not suitable in many cases.
One approach to reducing this cost is to only use as much computation as needed for the particular input.

Adaptive Computation Time (ACT)~\cite{graves2016adaptive} is a recently proposed mechanism that adjusts the computational depth of deep models: the harder the object is, the more iterations it is processed for.
This mechanism is end-to-end trainable, problem-agnostic and does not require an explicit supervision for the number of computational iterations.
It has been applied to recurrent networks for the problems of text modelling~\cite{graves2016adaptive} and reasoning~\cite{neumann2016learning}.
Spatially Adaptive Computation Time (SACT)~\cite{figurnov2017spatially} applies the ACT mechanism to the spatial positions of Residual Networks \cite{he2016identity}, a popular convolutional neural network model.
This results in computational savings and interpretable computation time maps that highlight the regions of the image that the network considers relevant to the task at hand.

In this paper, we introduce Probabilistic Adaptive Computation Time (PACT), a probabilistic model with discrete latent variables that specify the number of iterations to execute.
We define a prior on the latent variables that encodes the desired trade-off between speed and accuracy.
Then, we perform amortized maximum a posteriori (MAP) inference to find the proper amount of computation for a given object.
The ACT mechanism can be seen as an ad-hoc relaxation of the PACT model with a specific prior distribution.
A significant downside of the ACT relaxation is that it provides a discontinuous objective.
Since reparameterization trick is only valid for continuous objectives, ACT cannot be incorporated into stochastic models trained with reparameterization, such as variational autoencoder~\cite{kingma2014auto}.

We extend variational optimization~\cite{staines2012variational,staines2013optimization}, a method for MAP inference, to handle intractable expectations using REINFORCE or reparameterization trick.
For discrete latent variables, we propose to apply the Concrete relaxation~\cite{maddison2017concrete,jang2017categorical} and then perform the reparameterization.
We call the obtained method stochastic variational optimization and apply it to the PACT model.
Evaluation on ResNets shows that training using the relaxation outperforms the REINFORCE based method and matches the performance of the heuristic ACT.
We show that the relaxation allows to train the model with up to $1344$ discrete latent variables.
Additionally, the models trained with the proposed relaxation can be evaluated with a simple deterministic approach that reduces the memory consumption, compared to ACT.
Evaluation of the ACT models in the same manner decreases the performance.

\section{Background}

\paragraph{Notation.}
Let $\E_{q(z)} f(z)$ be the expectation of a function $f(z)$ over a probability distribution $q(z)$, $\sigma(z) = \tfrac{1}{1 + \exp(-z)}$ the sigmoid function, $\sigma^{-1}(z) = \log\tfrac{z}{1-z}$ the logit function, $[\mathrm{cond}]$ the step-function that is equal to $1$ if $\mathrm{cond}$ is true and $0$ otherwise.
Also, let $z_{<k}$ be a shorthand notation for $z_1, \dots, z_{k-1}$.

\subsection{Variational Optimization}

Variational optimization~\cite{staines2012variational,staines2013optimization} is a method for maximization of a function $f(z)$ of an argument $z$.
This argument can be either continuous or discrete.
To apply variational optimization, we choose an auxiliary parametric probability distribution over the arguments values $q_\phi(z)$.
The following lower bound on the optimal value holds for any distribution $q_\phi(z)$:
\begin{equation}
    L(\phi) = \E_{q_\phi(z)} f(z) \leq \E_{q_\phi(z)} \max_{z} f(z) = \max_{z} f(z).
    \label{eqn:varopt-bound}
\end{equation}
Suppose that the parametric family of distributions $q_\phi(z)$ can model arbitrary delta-functions.
Then, the bound is tight and the optimum is achieved when $q_\phi(z) = \delta(z - z^*)$, where $f(z^*) = \max_z f(z)$.

Let us assume that the density $q_\phi(z)$ is a smooth function of $\phi$.
Then, $L(\phi)$ is a smooth function.
Variational optimization further assumes that the expectation in $L(\phi)$ is tractable and maximizes $L(\phi)$ with a gradient-based method.
However, it is not applicable when the expectation is intractable.
We address this limitation in sec.~\ref{sec:stochastic-variational-optimization}.

\subsection{Variational Optimization for Probabilistic Models}

Consider a discriminative probabilistic model with latent variables $p(y, z | x) = p(y | x, z) p(z)$, where $x$ is the object, $y$ is the target label and $z$ is the latent variable.
The prior $p(z)$ encodes our preference for the values of $z$.
The maximum a posteriori (MAP) inference problem is to find $z^*$ that maximizes the density of the posterior distribution $p(z | x, y) = \frac{p(y, z | x)}{p(y | x)}$.
During training time, we know both $x$ and $y$, while during testing time we only have $x$ and would like to find the distribution $y$.
Therefore, we search for $z^*$ in a parametric form that only depends on $x$, so that we can use it during the test time.
This can be achieved by performing variational optimization with an auxiliary distribution $q_\phi(z | x)$:
\begin{equation}
    L_{\mathrm{MAP}}(\phi) = \E_{q_\phi(z | x)} ( \log p(y |  x, z) + \log p(z) ).
    \label{eqn:vo-map}
\end{equation}
For training, we plug in the ground-truth label $y$ and optimize $L_{\mathrm{MAP}}(\phi)$.
During testing, we sample $z \sim q_\phi(z | x)$ and obtain the distribution over the labels $p(y |  x, z)$.

Let us analyze a special case of this approach that has been extensively used in attention models literature~\cite{sohn2015learning,ba2015multiple,ba2015learning,xu2015show,li2017dynamic}.
Consider a probabilistic model $p_\phi(y, z | x) = p(y | x, z) p_\phi(z | x)$ with a learnable prior.
We can use the prior $p_\phi(z | x)$ as the approximate posterior in variational inference.
The corresponding evidence lower bound is
\begin{equation}
    L_{\mathrm{ML}}(\phi) = \E_{p_\phi(z | x)} \log p(y | x, z) \leq \log p_\phi(y | x).
\end{equation}
Renaming $p_\phi(z | x)$ into $q_\phi(z | x)$, we recognize the objective \eqref{eqn:vo-map}, where the prior distribution is uniform, $p(z) \propto 1$ (for a continuous latent variable on unbounded domain, this prior is improper).
Applying the inequality~\eqref{eqn:varopt-bound}, we have $L_{\mathrm{ML}}(\phi) \leq \max_z \log p(y | x, z)$.
Therefore, optimization of $L_{\mathrm{ML}}(\phi)$ corresponds to maximum likelihood inference of the latent variables.
On the other hand, the bound~\eqref{eqn:vo-map} allows to incorporate an explicit prior distribution over the latent variables and perform MAP inference.
This is a crucial requirement for the models such as the one proposed in the paper that provide an explicit prior distribution.

The objective~\eqref{eqn:vo-map} can also be seen as evidence lower bound on the marginal likelihood minus the entropy term.
Indeed, adding the entropy of $q_\phi(z | x)$ to the eqn. \eqref{eqn:vo-map} yields
\begin{equation}
    \E_{q_\phi(z | x)} \log \tfrac{p(y |  x, z) p(z)}{q_\phi(z | x)} \leq \log p(y | x).
\end{equation}
Unlike MAP inference, variational inference provides a distribution over the latent variable.
In our case, this is undesirable since we are interested in the single ``best'' value for the latent variables at the test time.
To obtain a single value of the variables for evaluation, we could choose a maximum of the approximate posterior.
However, this would introduce a gap between the train- and test-time behavior of the model.

\subsection{Concrete Distribution and Reparametrization}
\label{sec:concrete}

Suppose that we would like to stochastically optimize parameters $\phi$ of an intractable expectation $\E_{q_\phi(z)} f(z)$, where $f(z)$ is smooth.
The \textit{reparametrization trick} \cite{kingma2014auto,titsias2014doubly} allows for this, provided that the distribution $q_\phi(z)$ can be reparametrized, \ie we can sample $z \sim q_\phi(z)$ as follows:
\begin{equation}
    \varepsilon \sim q(\varepsilon),\ z = g(\varepsilon, \phi),
\end{equation}
where $g(\varepsilon, \phi)$ is smooth w.r.t. $\varepsilon$ and $\phi$.
Then, applying the chain rule we have:
\begin{equation}
    \nabla_\phi \E_{q_\phi(z)} f(z) = \E_{q(\varepsilon)} f'(g(\varepsilon, \phi)) \nabla_\phi g(\varepsilon, \phi).
\end{equation}
This expectation can be approximated using Monte-Carlo sampling.
The reparameterization trick is most commonly used for Normal distribution.
If $z \sim \operatorname{Normal}(\mu, \sigma^2)$, then $q(\varepsilon) = \operatorname{Normal}(0, 1)$ and $g(\varepsilon, \phi) = \mu + \varepsilon \sigma$.

Unfortunately, the reparameterization trick cannot be directly applied to discrete random variables, since the corresponding function $g(\varepsilon, \phi)$ is a non-smooth step function.
However, it is possible to \textit{relax} a discrete random variable so that the relaxation becomes reparameterizable.

The Concrete distribution \cite{maddison2017concrete,jang2017categorical} is a continuous reparameterizable relaxation of a discrete random variable.
For the purposes of this paper, we only consider relaxation of Bernoulli (binary) discrete random variables.
Consider a random variable $v \sim \operatorname{Bernoulli}(\gamma)$, where $p(v = 1) = \gamma \in (0, 1)$.
We introduce a temperature parameter $\lambda > 0$.
The relaxed random variable $\hat{v} \sim \operatorname{RelaxedBernoulli}(\gamma; \lambda)$ is defined via the following sampling procedure:
\begin{align}
    \varepsilon \sim \operatorname{Uniform}(0, &1),\ l = \sigma^{-1}(\gamma) + \sigma^{-1}(\varepsilon), \\
    &\hat{v} = \sigma\left(\frac{l}{\lambda}\right).
\end{align}

The $\operatorname{RelaxedBernoulli}$ distribution has several useful properties \cite{maddison2017concrete}.
First, the probability to be greater than 0.5 is equal for $\operatorname{Bernoulli}$ and $\operatorname{RelaxedBernoulli}$ random variables.
However, the mean value of $\operatorname{RelaxedBernoulli}$ is, in general, \textit{not} equal to $\gamma$.
For $\lambda \to 0$, the distribution of $\hat{v}$ approaches $\operatorname{Bernoulli}(\gamma)$.
Next, for $\lambda \leq 1$ the density $p(\hat{v})$ does not have modes in the interior of the $(0, 1)$ range.
As a result, the samples are typically close to either zero or one, which makes the relaxation work well for our purposes.
Importantly for us, when $\gamma \to 0$ or $\gamma \to 1$, the distribution of $\operatorname{RelaxedBernoulli}$ approaches a delta-function at 0 or 1, respectively.
This means that for extreme values of probability, the gap between the relaxed and non-relaxed distributions vanishes, regardless of the temperature $\lambda$.

\section{Stochastic Variational Optimization}
\label{sec:stochastic-variational-optimization}

Consider the variational optimization objective $L(\phi) = \E_{q_\phi (z)} f(z)$, where $z$ is a latent variable.
Stochastic variational optimization estimates the gradient $\nabla_\phi L(\phi)$ stochastically, even when the expectation is intractable.
First, we consider the case of a reparameterizable distribution, and then cover the case of discrete distributions.

If the distribution $q_\phi (z)$ is reparameterizable, \eg is a Normal distribution, we can perform reparameterization trick and calculate the stochastic gradients directly.
We then apply stochastic gradient optimization methods, resulting in stochastic variational optimization of the objective.

Now, we switch to the case where $z$ is discrete.
One popular method for this type of problems is REINFORCE~\cite{williams1992simple} training rule:
\begin{equation}
    \nabla_\phi L(\phi) = \E_{q_\phi(z)} ( f(z) - c ) \nabla_\phi \log q_\phi(z),
\end{equation}
where $c$ is a scalar baseline.
The expectation can be approximated by Monte-Carlo sampling.
Although this procedure provides unbiased gradients, the estimate often has an impractically high variance.

We propose to apply Concrete relaxation to the proposal distribution and then use the reparameterization trick.
This results in lower-variance gradients at the cost of a bias.
Assume that $z \in \{0,1\}^d$.
Let's decompose the proposal distribution using the chain rule, $q_\phi (z) = \prod_{i=1}^d q_\phi (z_i | z_{<i})$ (this sidesteps enumeration of all the $2^d$ configurations of $z$ during sampling).
We make two assumptions: (1) $f(z)$ is defined and smooth for $z \in [0, 1]^d$; (2) each factor $q_\phi (z_i | z_{<i}),\ i > 1$ is defined and smooth for $z_{<i} \in [0,1]^{i - 1}$.
Then, we can apply the Concrete relaxation with temperature $\lambda > 0$ to each factor (the hat denotes relaxation):
\begin{equation}
    q_{\phi, \lambda} (\hat{z}) = \prod_{i=1}^d q_{\phi,\lambda} (\hat{z}_i | \hat{z}_{<i}).
\end{equation}
The relaxed objective has the form
\begin{equation}
    \hat{L}_\lambda(\phi) = \E_{q_{\phi,\lambda}(\hat{z})} f(\hat{z}).
\end{equation}
This objective can now be stochastically optimized using the reparameterization trick.

If all the probabilities in the relaxed distribution approach extreme values (0 or 1), the relaxed distribution approaches the non-relaxed one, for any temperature $\lambda$.
In this case, the value of the relaxed objective $\hat{L}_\lambda (\phi)$ approaches the value of the original objective $L (\phi)$.

\section{Probabilistic Adaptive Computation Time}

First, we introduce \textit{adaptive computation block}.
It is a computation module that chooses the number of iterations depending on the input.
Depending on the specific type of the latent variables, we obtain a \emph{discrete}, \emph{thresholded} or \emph{relaxed} block.
Importantly, the blocks are compatible in the sense that one can train a model with one type of block and then switch to another during evaluation.
Then, we present a probabilistic model that incorporates the number of iterations as a latent variable into a discriminative model.
The prior on the latent variable favors using less iterations.
Finally, we perform MAP inference over the number of iterations via stochastic variational optimization.

\textbf{Discrete adaptive computation block} performs $z \in \{1, \dots, L\}$ iterations of computation, where $z$ is a discrete latent variable.
Let us assume that the $l$-th iteration outputs a value $u^l$ (we use upper subscripts to index the iterations in a block), and that all $u^1, \dots, u^L$ have the same shape.
The output of the block is $u^z$, the output of the $z$-th iteration.
To perform optimization over the discrete latent variable $z$, we introduce a distribution $q_\phi(z)$ with parameters $\phi$.
Denote $z^l = [z = l]$ the \textit{halting unit} of the block: when it is equal to one, the computation is halted.
The two desiderata for $q_\phi(z)$ are: (1) the probability of halting at the $l$-th step should depend on $u^l$; (2) it should be possible to sample $z^l$ after only executing the first $l$ iterations.

To satisfy the first property, we introduce a \textit{halting probability} for every iteration:
\begin{equation}
    h^l = H_{\phi}^l (u^l),\ l = 1, \dots, (L - 1),\ h^L = 1.
\end{equation}
For the second property, we define the following sampling procedure for the distribution $q_\phi(z)$:
\begin{align}
    \xi^l &\sim \operatorname{Bernoulli}(h^l),\ l = 1 \dots L-1,\ \xi^L = 1, \label{eqn:bernoulli-sample} \\
    z^l &= \xi^l \prod_{i=1}^{l-1} (1-\xi^i),\ l = 1 \dots L.
    \label{eqn:q-sampling}
\end{align}
The vector $(z^1, \dots, z^L)$ is a one-hot representation of the discrete $L$-ary latent variable $z$.
We reparameterize $z$ via $(L-1)$ Bernoulli latent variables $(\xi^1, \dots, \xi^{L-1})$.
The distribution of $z$ can be obtained by taking an expectation over the independent random variables $\xi^l$:
\begin{equation}
    q_\phi(z^l = 1) = q_\phi(z = l) = h^l \prod_{i=1}^{l-1} (1 - h^i).
    \label{eqn:proba}
\end{equation}

\begin{figure}
    \centering
    \includegraphics[width=0.9\linewidth]{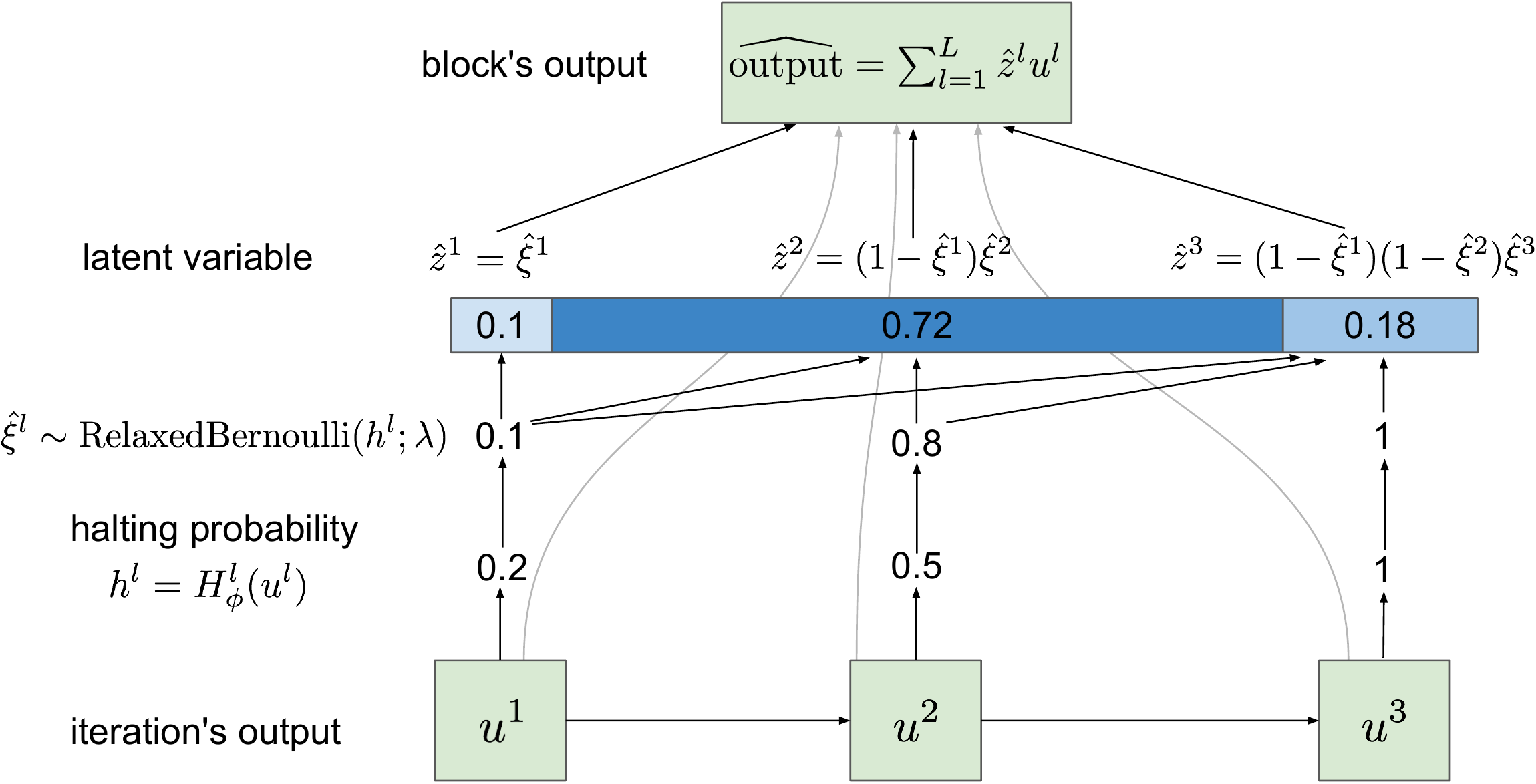}
    \caption{Relaxed adaptive computation block.}
    \label{fig:relaxed-block}
\end{figure}

\textbf{Thresholded adaptive computation block} is a deterministic version of the (stochastic) discrete adaptive computation block.
Since we perform MAP inference over the latent variables, we expect the halting probabilities $h^l$ to be sufficiently close to either zero or one.
Therefore, during evaluation we can replace sampling~\eqref{eqn:bernoulli-sample} with thresholding of the halting probabilities:
\begin{equation}
    \xi^l = [h^l > 0.5].
    \label{eqn:thresholded}
\end{equation}
The advantage of this block is an extremely simple implementation: stop as soon as the halting probability exceeds $0.5$.

\textbf{Relaxed adaptive computation block} is obtained form discrete adaptive computation block by replacing the $\operatorname{Bernoulli}$ random variables with $\operatorname{RelaxedBernoulli}$ variables.
We denote the relaxed variables with a hat and define the temperature of the relaxation $\lambda > 0$.
Sampling  the vector $\hat{z} = (\hat{z}^1, \dots, \hat{z}^L)$ from $q_{\phi, \lambda}(\hat{z})$ proceeds as follows:
\begin{align}
    &\hat{\xi}^l \sim \operatorname{RelaxedBernoulli}(h^l; \lambda),\ l = 1 \dots L-1,\\
    &\hat{\xi}^L = 1, \hat{z}^l = \hat{\xi}^l \prod_{i=1}^{l-1} (1-\hat{\xi}^i),\ l = 1 \dots L.
\end{align}
The vector $\hat{z}$ is no longer one-hot.
However, since it is produced by a stick-breaking procedure, it forms a discrete probability distribution over the iterations that we call the \emph{halting distribution}.
Finally, we define the output of the relaxed adaptive computation block as an expectation of the iteration outputs w.r.t. the halting distribution $\hat{z}$:
\begin{equation}
    \widehat{\mathrm{output}} = \sum_{l=1}^L \hat{z}^l u^l.
\end{equation}
The whole procedure is illustrated on fig.~\ref{fig:relaxed-block}.

\textbf{Probabilistic model}.
Consider a discriminative model with a likelihood $p_\theta(y | x)$ of the target label $y$ given an object $x$ (for simplicity of notation, we consider just one object), parameterized by $\theta$.
This model can be a deep network for classification or regression problem.
In many cases we prefer that the model make the prediction as quickly as possible.
Assume that we have incorporated $K$ adaptive computation blocks into the likelihood with the corresponding latent variables (number of computation iterations) $\mathbf{z} = (z_1, \dots, z_K)$.
Also, denote the maximum number of iterations in the $k$-th block as $L_k$.

We now discuss the prior distribution $p(\mathbf{z})$ that encodes the preference for less iterations.
For simplicity, we assume that it factorizes over the blocks, $p(\mathbf{z}) = \prod_{k=1}^K p(z_k)$.
The prior for each block $p(z_k)$ is a discrete distribution over $L_k$ iterations.
To make our model directly comparable to ACT, we choose a prior distribution that provides the same log-linear penalty as the ACT model (up to a normalization constant), a truncated Geometric distribution.
We parameterize the Geometric distribution via a log-scale number of iterations penalty $\tau_k > 0$ (the canonical Geometric distribution probability for success $\alpha_k$ can be recovered as $\alpha_k = 1 - \exp(-\tau_k)$).
The prior distribution for a single block is
\begin{equation}
\small
\begin{split}
    &\operatorname{TruncatedGeometric}(z_k | \tau_k, L_k) \\
    &=\frac{\exp(\tau_k) - 1}{1 - \exp(-\tau_k L_k)} \exp(-\tau_k z_k),\ z_k \in \{1, \dots, L_k\}.
\end{split}
\end{equation}

Using the described prior, we obtain the following probabilistic model:
\begin{equation}
\begin{split}
    &p_\theta(y, \mathbf{z} | x) = p_\theta(y | x, \mathbf{z}) p(\mathbf{z}), \\
    &p(\mathbf{z}) = \prod_{k=1}^K \operatorname{TruncatedGeometric}(z_k | \tau_k, L_k) \\
    &= \left( \prod_{k=1}^K \frac{\exp(\tau_k) - 1}{1 - \exp(-\tau_k L_k)}\right) \exp\left(- \sum_{k=1}^K \tau_k z_k\right).
\end{split}
\end{equation}

We perform MAP inference of the latent variable $z$ by variational optimization with an auxiliary distribution
\begin{equation}
    q_\phi(\mathbf{z} | x) = \prod_{k=1}^K q_\phi(z_k | z_{<k}, x),
\end{equation}
where $q_\phi(z_k | z_{<k}, x)$ is defined via eqn. \eqref{eqn:q-sampling}.
The dependence on the input and the previous latent variables is via the inputs of the block.
We refer to this probabilistic model as \emph{discrete}.
The objective for maximization w.r.t. $\theta$ and $\phi$ is
\begin{equation}
\begin{split}
    &L(\theta, \phi) = \E_{q_\phi(\mathbf{z} | x)} \log p_\theta(y, \mathbf{z} | x) \\
    & = \E_{q_\phi(\mathbf{z} | x)} \left( \log p_\theta(y | \mathbf{z}, x) + \sum_{k=1}^K \log p(z_k)\right).
    \label{eqn:objective}
\end{split}
\end{equation}

To reduce the variance of the stochastic estimate of the objective, we analytically compute the expectation of the log-prior:
\begin{equation}
\begin{split}
    &\E_{q_\phi(\mathbf{z} | x)} \log p(z_k) \\
    &= -\tau_k \E_{q_\phi(z_{<k} | x)} \underbrace{\sum_{l=1}^{L_k} l\ h^l_k \prod_{i=1}^{l-1} (1-h^i_k)}_{N_k} + \mathrm{const}.
\end{split}
\end{equation}
Here $N_k$ is the expected number of iterations in the $k$-th block.
Ignoring the additive constant, we have
\begin{equation}
    L(\theta, \phi) = \E_{q_\phi(\mathbf{z} | x)} \left(\log p_\theta(y | \mathbf{z}, x) - \sum_{k=1}^K \tau_k N_k \right).
    \label{eqn:objective-2}
\end{equation}

The objective in eqn.~\eqref{eqn:objective-2} is intractable for deep models consisting of several stacked adaptive computation blocks, as the complexity of direct evaluation of the expectation grows exponentially in the number of blocks.
One heuristic is to replace the random variables $z_k$ with their expectations and optimize the probabilities directly.
However, this simple approach fails for deep networks as they learn to trick the objective by increasing the halting probability for the first iterations and decreasing it for the latter iterations, while significantly boosting the magnitude of the outputs for the latter iterations~\cite{graves2016adaptive}.
The prior term value reflects that few iterations were used, while the outputs of the blocks are dominated by the last iterations.

Instead, we stochastically optimize the objective~\eqref{eqn:objective-2}.
In sec.~\ref{sec:stochastic-variational-optimization} we proposed two approaches to do this, one using REINFORCE and another using relaxation.

In the first approach, we directly apply REINFORCE to the objective~\eqref{eqn:objective-2}, obtaining the following gradients w.r.t. $\phi$:
\begin{equation}
\label{eqn:reinforce}
\begin{split}
    &\nabla_\phi L(\theta, \phi) = \E_{q_\phi(\mathbf{z} | x)} \Bigl( (\log p_\theta(y | \mathbf{z}, x) - c) \\
    & \times \nabla_\phi \log q_\phi(\mathbf{z} | x) - \sum_{k=1}^K \tau_k \nabla_\phi N_k \Bigr),
\end{split}
\end{equation}
where $c$ is a scalar baseline.
The value $q_\phi(z | x)$ is defined by eqn.~\eqref{eqn:proba}.
Note that we have neglected the dependency of $N_k$ on $z_{<k}$ to reduce the variance of the gradients.

For the second approach, we replace every adaptive computation block with a relaxed counterpart, and the corresponding distribution $q_\phi(z)$ with the relaxed distribution $q_{\phi, \lambda}(\hat{z})$.
This \textit{relaxed} model has an objective that can be optimized via the reparameterization trick:
\begin{equation}
    \hat{L}_\lambda(\theta, \phi) = \E_{q_{\phi, \lambda}(\hat{\mathbf{z}} | x)} \left( \log p_\theta(y | \hat{\mathbf{z}}, x) - \sum_{k=1}^K \tau_k N_k \right).
    \label{eqn:objective-3}
\end{equation}

In the supplementary we present the algorithms for PACT in Discrete, Thresholded and Relaxed modes.

\subsection{Application: Probabilistic Spatially Adaptive Computation Time for Residual Networks}

Residual network (ResNet) \cite{he2016deep,he2016identity} is a deep convolutional neural network architecture that has been successfully applied to many computer vision problems \cite{dai2016rfcn,chen2016deeplab}.
We describe ResNet-32 and ResNet-110 models for CIFAR image classification dataset \cite{krizhevsky2009cifar}.
They contain three stacked \textit{blocks}, each consisting of several \textit{residual units} (5 for ResNet-32 and 18 for ResNet-110).
The computational iteration of a ResNet is a residual unit of the form $F^l_k(u^{l-1}_k) = u^{l-1}_k + f^l_k(u^{l-1}_k)$, where $f^l_k$ is a sub-network consisting of two convolutional layers.
$u^0_k$ is the output of the previous block of residual units.
The outputs of the residual units in each block have the same size.
The first units in the second and third blocks are applied with stride 2 to perform spatial downsampling, while also increasing the number of output channels by a factor of two.
Thus, the spatial dimensions of the first block are $32\times32$ (same as the size of CIFAR-10 images), the second block $16\times16$ and the third block $8\times8$.
In this way, the amount of computation for every residual unit is roughly constant.
The outputs of the last block are passed through a global average pooling and linear layers to obtain the class probabilities logits.

SACT \cite{figurnov2017spatially} applies the ACT mechanism to every spatial position of every residual network block.
Likewise, we apply an adaptive computation block to every spatial position of every residual network block.
We call the obtained model \emph{PSACT}, probabilistic spatially adaptive computation time.
The corresponding latent variable is $z_{k,ij}$ where $k$ is the number of residual network block and $ij$ is the spatial position.
The halting probability map is computed as $H^l_k(u) = \sigma (\widetilde{W}^l_k \ast u + W^l_k \operatorname{pool}(u) + b^l_k)$, where $\ast$ is $3\times3$ convolution and $\operatorname{pool}$ is global average pooling.
The computation time penalty for a block is chosen to be $\frac{\tau}{HW}$, where $\tau$ is a global computation time penalty and $H$ and $W$ are the height and width of the ResNet block.

In order to impute the non-computed intermediate values, we redefine the residual unit as
\begin{equation}
    F^l_k(u^{l-1}_k) = u^{l-1}_k + f^l_k (u^{l-1}_k) \cdot a(\xi^{<l}_k),
\end{equation}
where $a(\xi^{<l}_k)$ is an \textit{active positions mask}.
For the discrete model, we choose $a(\xi^{<l}_k) = \prod_{t=1}^{l-1} (1 - \xi^t_k)$, with the operation performed element-wise.
Thus, if the position is no longer evaluated (hence, $z_k < l$), the value is zero and we simply carry the features from the previous iteration.
Otherwise, the value is one.
For the relaxed model, we use $\hat{a}(\hat{\xi}^{<l}_k) = r \cdot [r > \delta],\ r = \prod_{t=1}^{l-1} (1 - \hat{\xi}^t_k)$, where $\delta > 0$ is a scalar hyperparameter.
By clipping the values of $r$, we obtain strict zeros and can skip computing the corresponding values during the training time.
We have verified that setting $\delta$ to zero gives similar results, although without a possibility of computation savings during training.

\subsection{Application: Probabilistic Adaptive Computation Time for Recurrent Neural Networks}

We can also apply the proposed model to dynamically vary the amount of computation in Recurrent Neural Networks, such as Long Short-Term Memory networks (LSTMs) \cite{hochreiter1997long}.
Let us denote the input sequence $\mathbf{x} = (x_1, \dots, x_T)$, where $T$ is the number of timesteps.
An adaptive computation block is associated with each timestep.
Therefore, each timestep is processed for an adaptive number of iterations.
We can use the same computation time penalty $\tau$ for all iterations.
The computation iteration consists of applying the RNN's transition function to obtain the new state of the RNN: $u^l_k = F_\theta(x_k, [l=1], u^{l-1}_k)$.
Here $u^0_k$ is the output state from the previous block/timestep.
The binary input feature $[l = 1]$ allows the network to detect the beginning of a new timestep.
The halting probability is computed as $h^l_k = H_{\phi}(u^l_k) = \sigma (W u^l_k + b)$.
The output state of a block is used as an input state for the next block and as features for predicting the emission values for the timestep.

\section{Related work}

Adaptive Computation Time (ACT) mechanism \cite{graves2016adaptive} can be seen as a heuristic deterministic relaxation of our PACT model.
Specifically, ACT transforms the halting probabilities $(h^1, \dots, h^L)$ into the halting distribution $(\hat{z}^1, \dots, \hat{z}^L)$ as follows:
\begin{align}
N &= \min \Big\{ n \in \{1 \dots L\}: \sum_{l=1}^{n} h^l \geq 0.99 \Big\}, \label{eqn:act-n}\\
R &= 1 - \sum_{l=1}^{N-1} h^l, \ \hat{z}^l =
            \begin{cases}
            h^l &\text{if } l < N, \\
            R &\text{if } l = N, \\
            0 &\text{if } l > N.
            \end{cases}
\end{align}
Since the halting distribution is not one-hot, additional memory is required to maintain the output $\sum_{l=1}^L \hat{z}^l u^l$ during evaluation (an algorithm is presented in the supplementary).
In discrete and thresholded PACT models, the halting distribution is one-hot and this memory can be saved.

\begin{figure}
    \centering
    \includegraphics[width=0.7\linewidth]{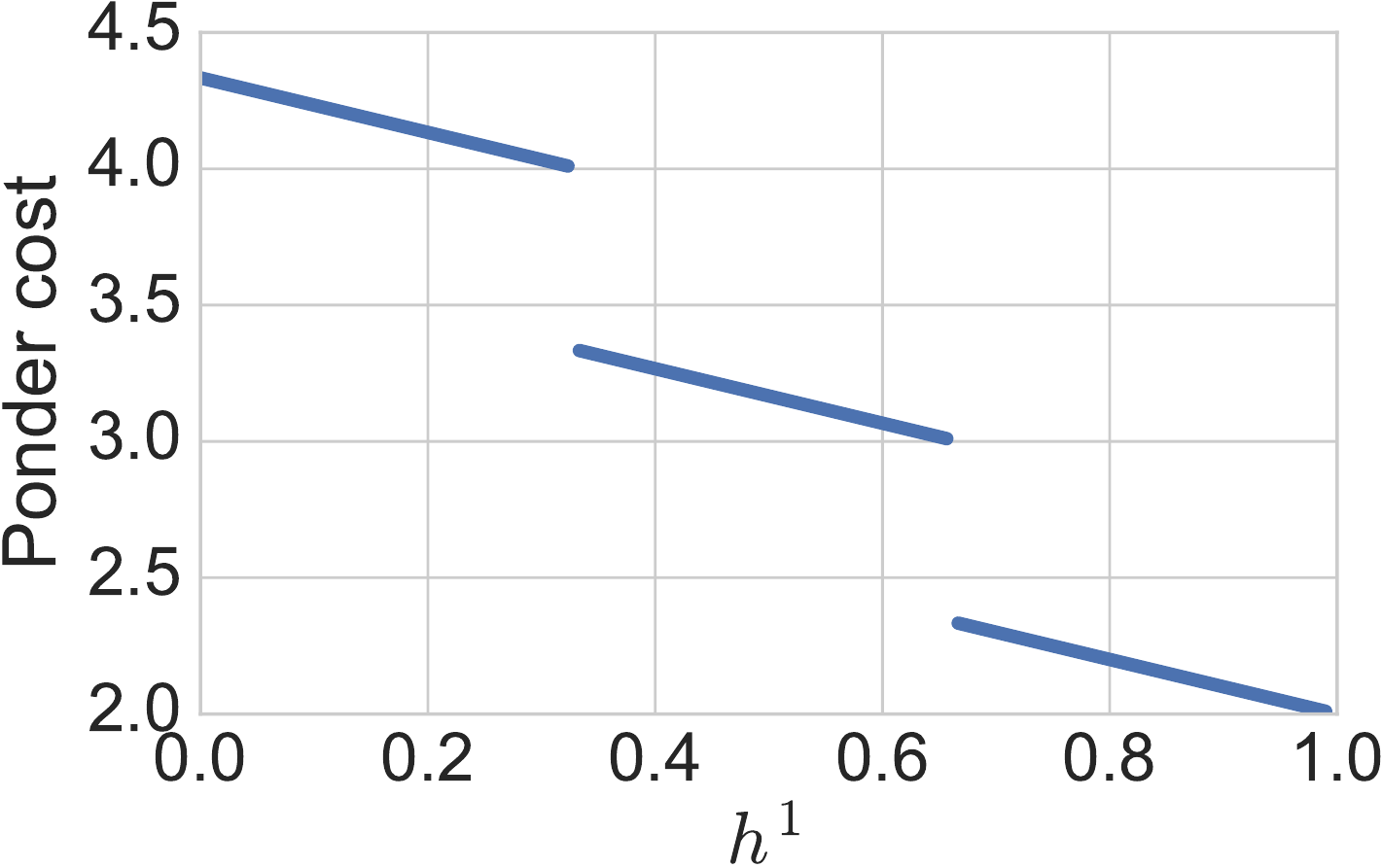}
    \caption{Ponder cost $\rho$ is a discontinuous function of the halting probability $h^1$. Here $h^2 = h^3 = h^4 = \nicefrac{1}{3}$.}
    \label{fig:ponder-cost-discontinuous}
\end{figure}

The stopping time $N$ has zero gradients almost everywhere.
In order to optimize the stopping time, a differentiable upper bound, \textit{ponder cost}, $\rho = N + R$ is introduced.
Ponder cost is linear almost everywhere, but is a discontinuous function of the halting probabilities, with discontinuities arising in the configurations where $N$ changes the value, see fig. \ref{fig:ponder-cost-discontinuous}.
For instance, this means that ACT cannot be used with reparameterization trick that is only valid for continuous objectives.
The objective of ACT, for several adaptive computation blocks, is $\log p(y | \hat{\mathbf{z}}, x) - \sum_{k=1}^K \tau_k \rho_k$.

Let us summarize why the proposed PACT model is more principled than ACT.
First, the discrete PACT model straightforwardly defines the halting time as the iteration where the halting unit is fired.
On the other hand, ACT that uses an ad-hoc definition~\eqref{eqn:act-n}.
Second, PACT allows to directly minimize the expected halting time, while ACT minimizes the discontinuous ponder cost.

Several papers have explored using REINFORCE for adjusting the number of computation steps in neural networks using discrete latent variables: choosing the number of patches to process \cite{li2017dynamic}, determining the number of objects on a scene \cite{eslami2016attend}, dropping the unnecessary subsets of neurons in a fully-connected network \cite{bengio2016conditional}.
REINFORCE for discrete latent variables is also used for \textit{hard attention} methods \cite{mnih2014recurrent,ba2015multiple}.
Most of them use the same amount of computation for all inputs, although~\cite{li2017dynamic} explores dynamically adjusting the number of steps.
As we experimentally show, using Concrete relaxation dramatically simplifies training, compared to using REINFORCE.

Recently, \cite{jernite2017variable} proposed to only update a dynamically chosen subset of the hidden state of a recurrent network.
This can be seen as an alternative to ACT for recurrent neural networks.
However, it is still a heuristic mechanism requiring several tricks to train.

Two concurrent works explore adaptive dropping of residual units in ResNet models using Actor-Critic~\cite{wu2017blockdrop} and Gumbel-Softmax~\cite{veit2017convolutional}.
This can be seen as an adaptive version of stochastic depth~\cite{huang2016deep}.
In this paper, we propose a probabilistic view of ACT and SACT mechanisms.
The resulting method is generally applicable to sequential models, including ResNets and RNNs.

Our work follows a trend in machine learning of interpreting methods as approximate Bayesian procedures.
For example, in the field of topic modelling, Latent Dirichlet Allocation~\cite{blei2003latent} is a probabilistic counterpart of Latent Semantic Indexing~\cite{deerwester1990indexing}.
Recently, Dropout~\cite{srivastava2014dropout} has been interpreted as variational inference in a probabilistic model~\cite{kingma2015variational,gal2016dropout}.
This spurred the development of more innovative ways of using Dropout, \eg in RNNs~\cite{gal2016theoretically} and for sparsifying neural networks~\cite{molchanov2017variational}.
We hope that our paper will similarly open the way for various extensions of adaptive computation time.

\section{Experiments}

\begin{figure*}
\centering
\begin{subfigure}[t]{0.25\linewidth}
   \centering
   \includegraphics[width=\linewidth]{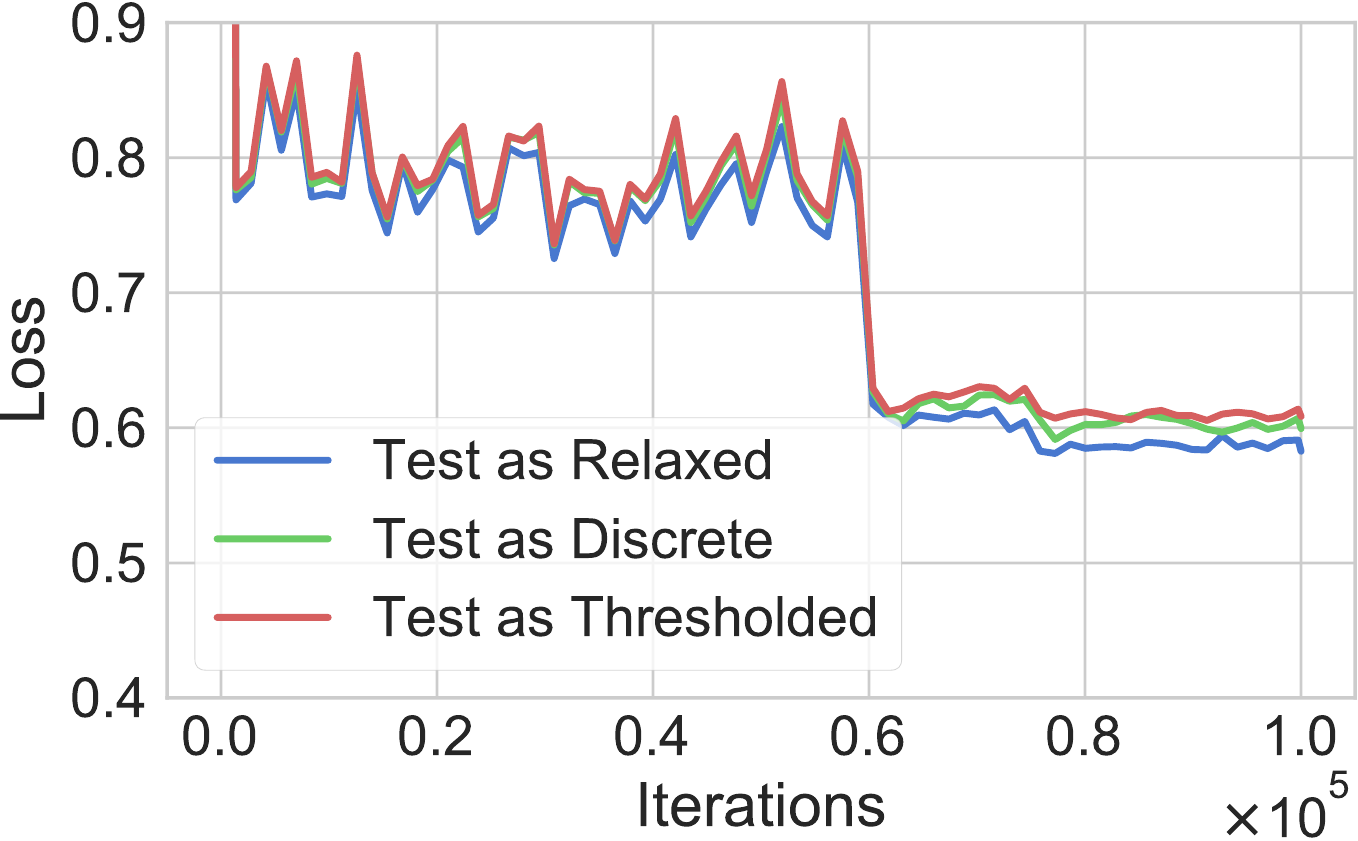}
\end{subfigure}
\quad
\begin{subfigure}[t]{0.25\linewidth}
   \centering
   \includegraphics[width=\linewidth]{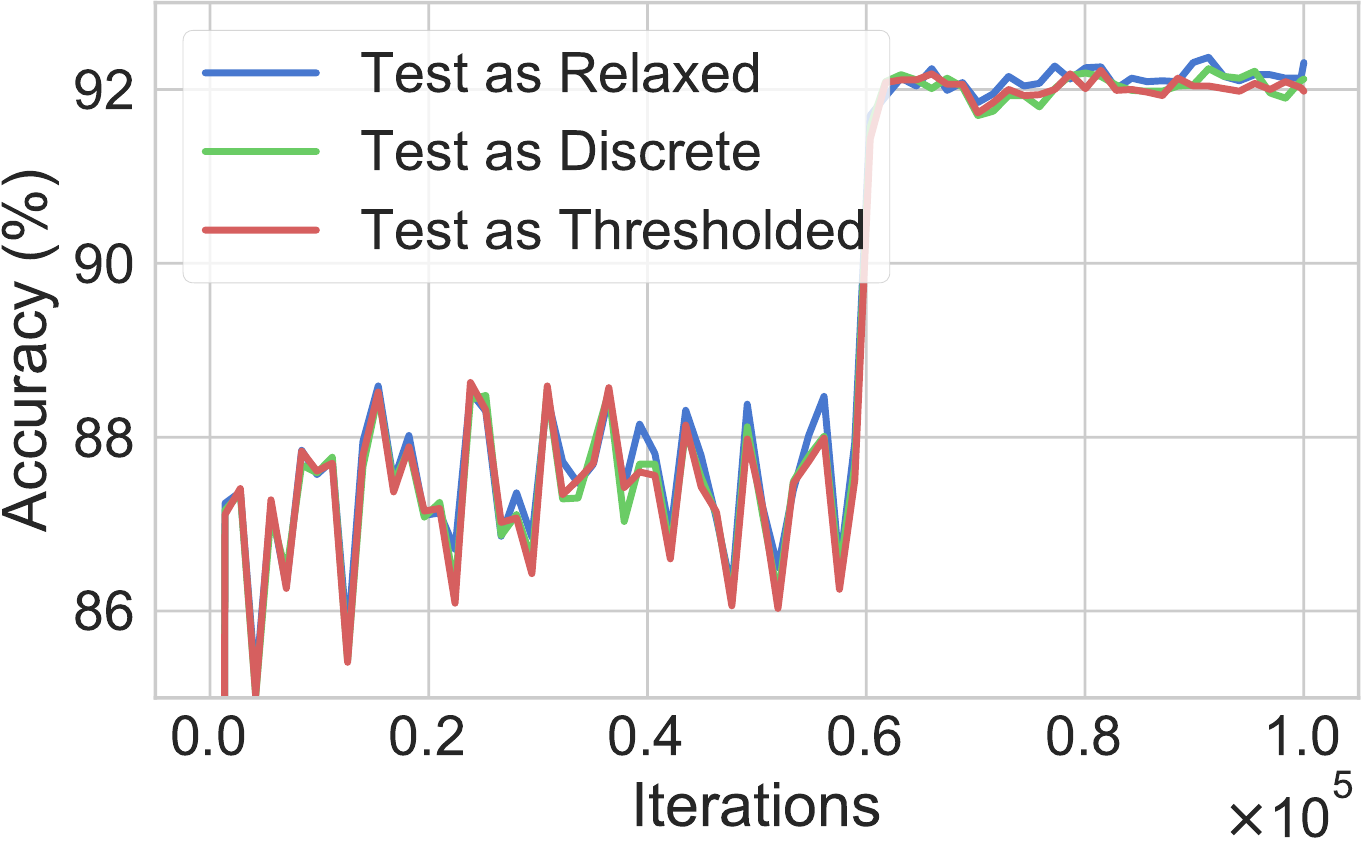}
\end{subfigure}
\quad
\begin{subfigure}[t]{0.25\linewidth}
   \centering
   \includegraphics[width=\linewidth]{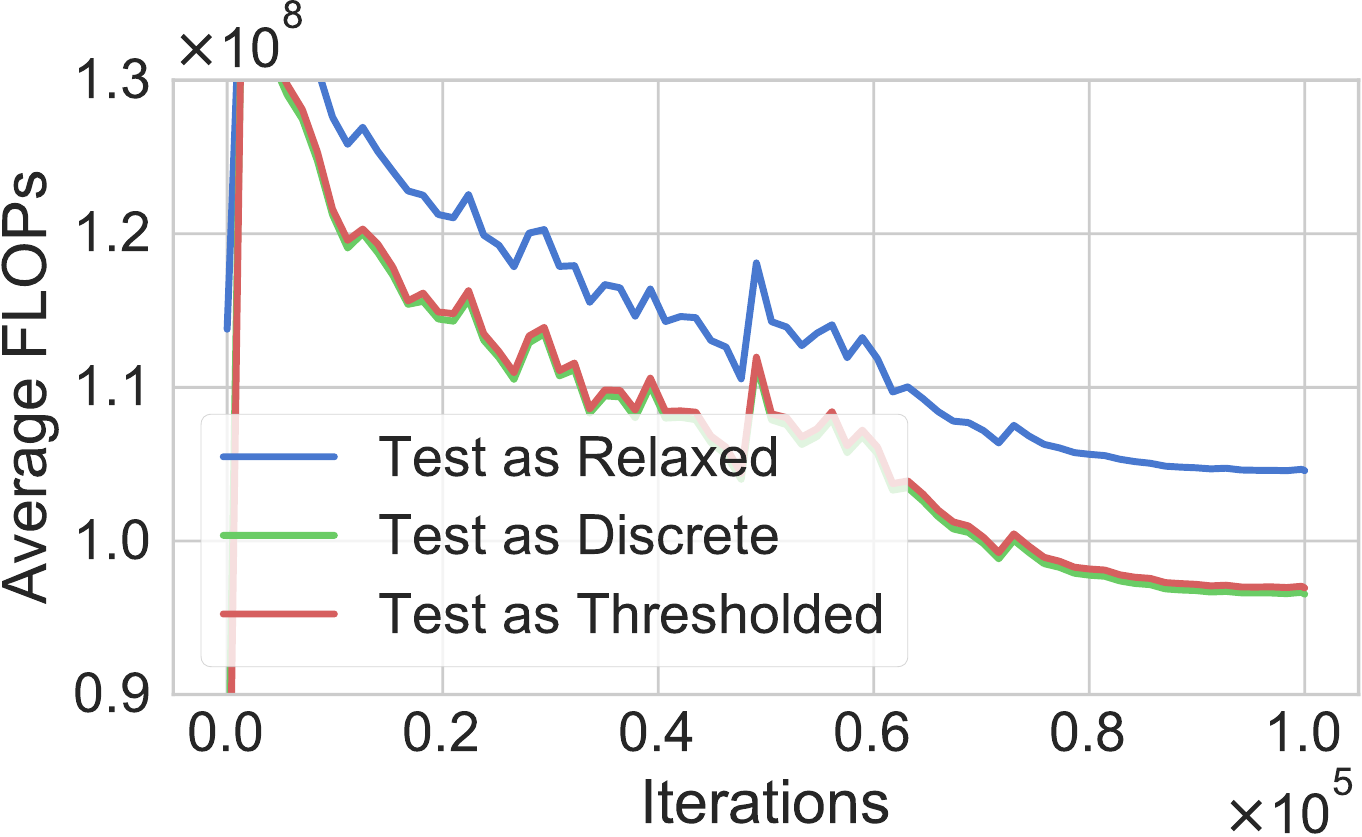}
\end{subfigure}
\caption{The parameters from a relaxed PSACT model (ResNet-32, $\tau=0.01$) for different training iterations are evaluated on the test set in Relaxed, Discrete and Thresholded models. The gap between the models is small throughout the training.}
\label{fig:relaxation}
\end{figure*}

\begin{figure}
\centering
\begin{subfigure}[t]{0.49\linewidth}
   \centering
   \includegraphics[width=\linewidth]{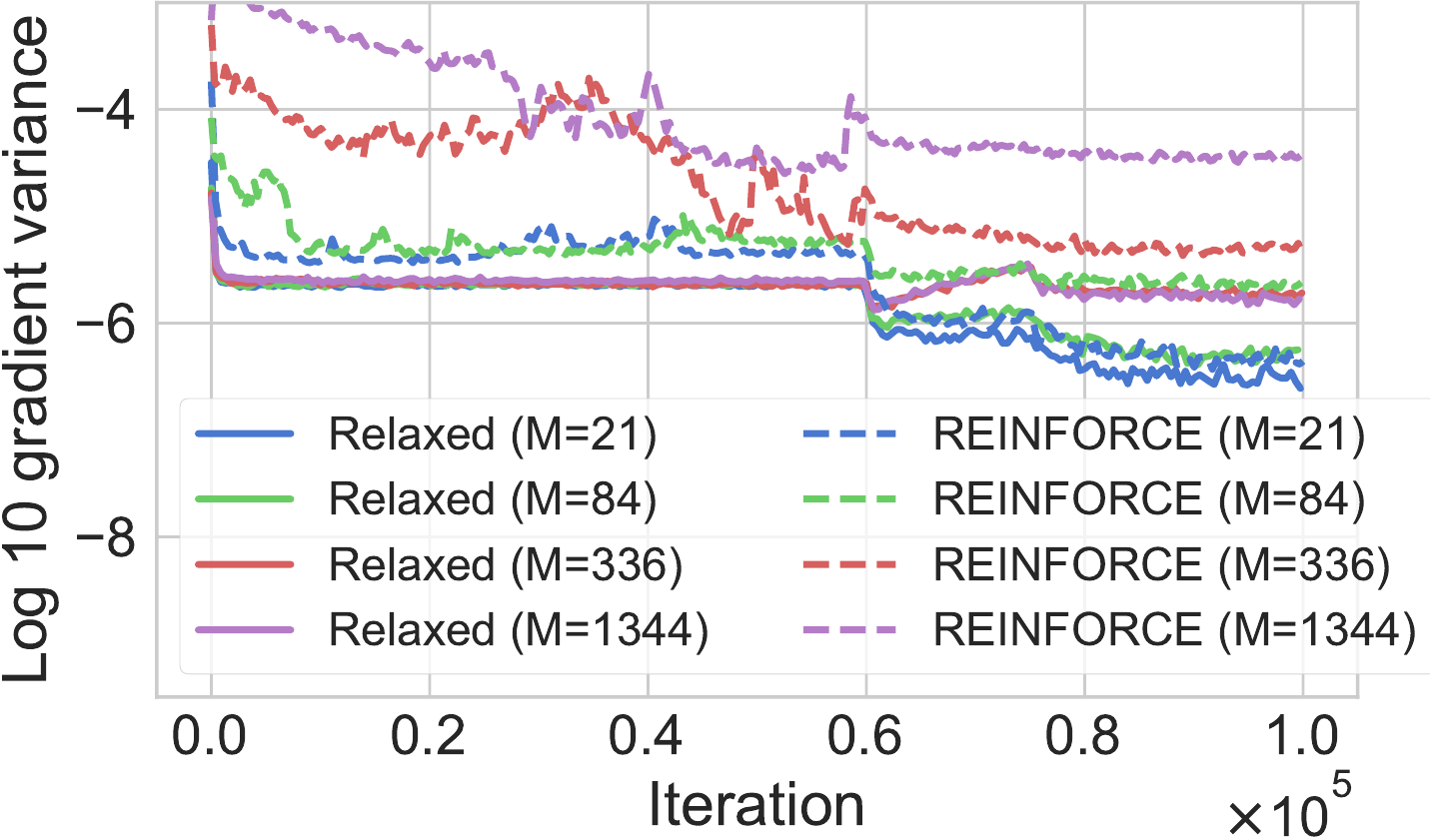}
\end{subfigure}
\begin{subfigure}[t]{0.49\linewidth}
   \centering
   \includegraphics[width=\linewidth]{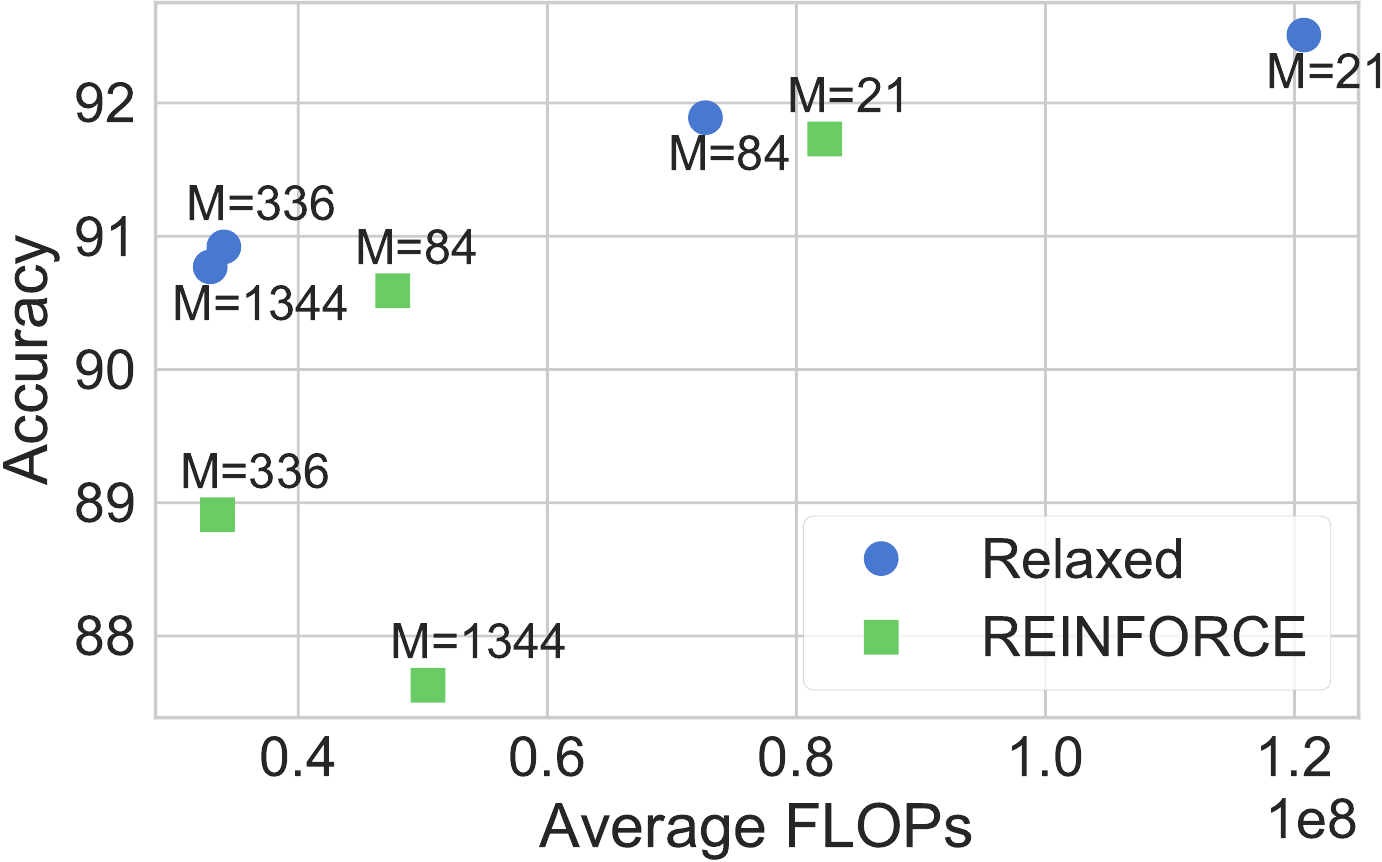}
\end{subfigure}
\caption{Training of relaxed PSACT model (ResNet-32, $\tau=0.1$) and training of discrete PACT model using REINFORCE, for varying number of the latent variables $M$. REINFORCE exhibits much higher variance of gradients and fails to reach a competitive accuracy for $M > 84$. \textbf{Left}: $\log_{10}$ of the parameters gradient variance as a function of the training iteration. \textbf{Right}: test FLOPs and accuracy at convergence (evaluation is performed in the discrete mode).}
\label{fig:reinforce}
\end{figure}

In the experimental evaluation we focus on PSACT model for ResNets, since it allows to adjust the number of latent variables by grouping the spatial positions.
First, we demonstrate that the relaxed model's parameters are compatible with the discrete and thresholded models.
Then, we compare training of the relaxed model to training of the discrete model with with REINFORCE, for varying number of latent variables.
Finally, we demonstrate that the relaxed PSACT model achieves close results to ACT.
We also verify that the parameters obtained by the relaxed model can be used in a thresholded model with extremely simple test-time behavior, and that it is not the case for SACT.

We consider pre-activation ResNets~\cite{he2016identity} with 32 and 110 convolutional layers.
We use CIFAR-10 image classification dataset~\cite{krizhevsky2009cifar}.
The training hyperparameters are provided in the supplementary.
Unless otherwise noted, PSACT is trained using the relaxed model and evaluated using the discrete model.
As a proxy to the potential time savings, we compute the number of floating point operations (FLOPs) required to evaluate the positions with non-zero values in the active positions mask, as done in \cite{figurnov2017spatially}.

In the first experiment, we train a relaxed PSACT model.
The obtained parameters are continuously evaluated on the test set in three models: relaxed (Concrete relaxation of the Bernoulli variables), discrete (discrete latent variables), and thresholded (deterministic latent variables).
The results on fig. \ref{fig:relaxation} show that the loss function and accuracy stay remarkably close for the three models.
Since the computation in relaxed model is stopped when $\prod_{i=1}^l (1 - \hat{\xi}^i_k) < \delta$, and $\hat{\xi}^i_k$ might take non-extreme values, the relaxed model requires more computation.

Next, we compare training of the relaxed model to training of the discrete model using REINFORCE.
We use an exponential moving average reward baseline with a decay factor of $0.99$.
We do not employ an input-dependent baseline to simplify the model, since the paper~\cite{mnih2014neural} finds small improvement from using it.
Additionally, for REINFORCE, we use Adam optimizer \cite{kingma2015adam} with initial learning rate of $10^{-3}$ (the decay schedule is kept the same), since SGD with momentum used in other experiments results in unstable training.

PSACT model for ResNet-32 has $M = 1344$ 5-ary categorical latent variables: one variable per $(32\cdot32 + 16\cdot16 + 8\cdot8)$ spatial positions.
To study the effect of the number of the latent variables on the training, we group the latent variables spatially.
Namely, in every ResNet block, we group the spatial positions into non-overlapping $n \times n$ patches, $n \in \{2, 4, 8\}$.
Within each patch, we average the logits of the halting probabilities and sample a single latent variable per patch.
The results presented on fig. \ref{fig:reinforce} show that REINFORCE has a much higher gradient variance.
For $M=1344$ latent variables, the difference is about two orders of magnitude.
REINFORCE achieves comparable results for $M=21$ and $M=84$ latent variables, but the accuracy quickly deteriorates when the number of latent units is increased.

\begin{figure}
\centering
\begin{subfigure}[t]{0.49\linewidth}
   \centering
   \includegraphics[width=\linewidth]{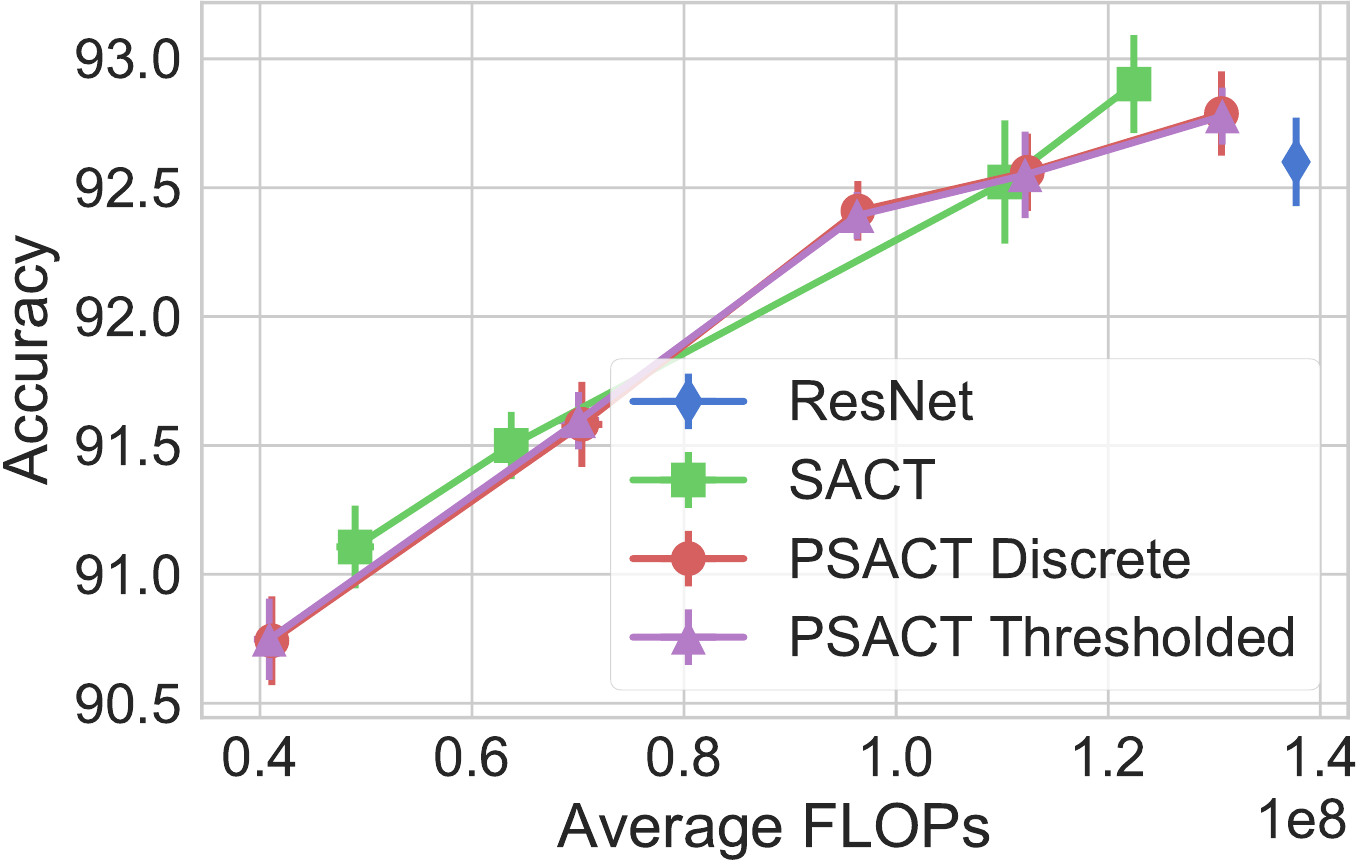}
\end{subfigure}
\begin{subfigure}[t]{0.49\linewidth}
   \centering
   \includegraphics[width=\linewidth]{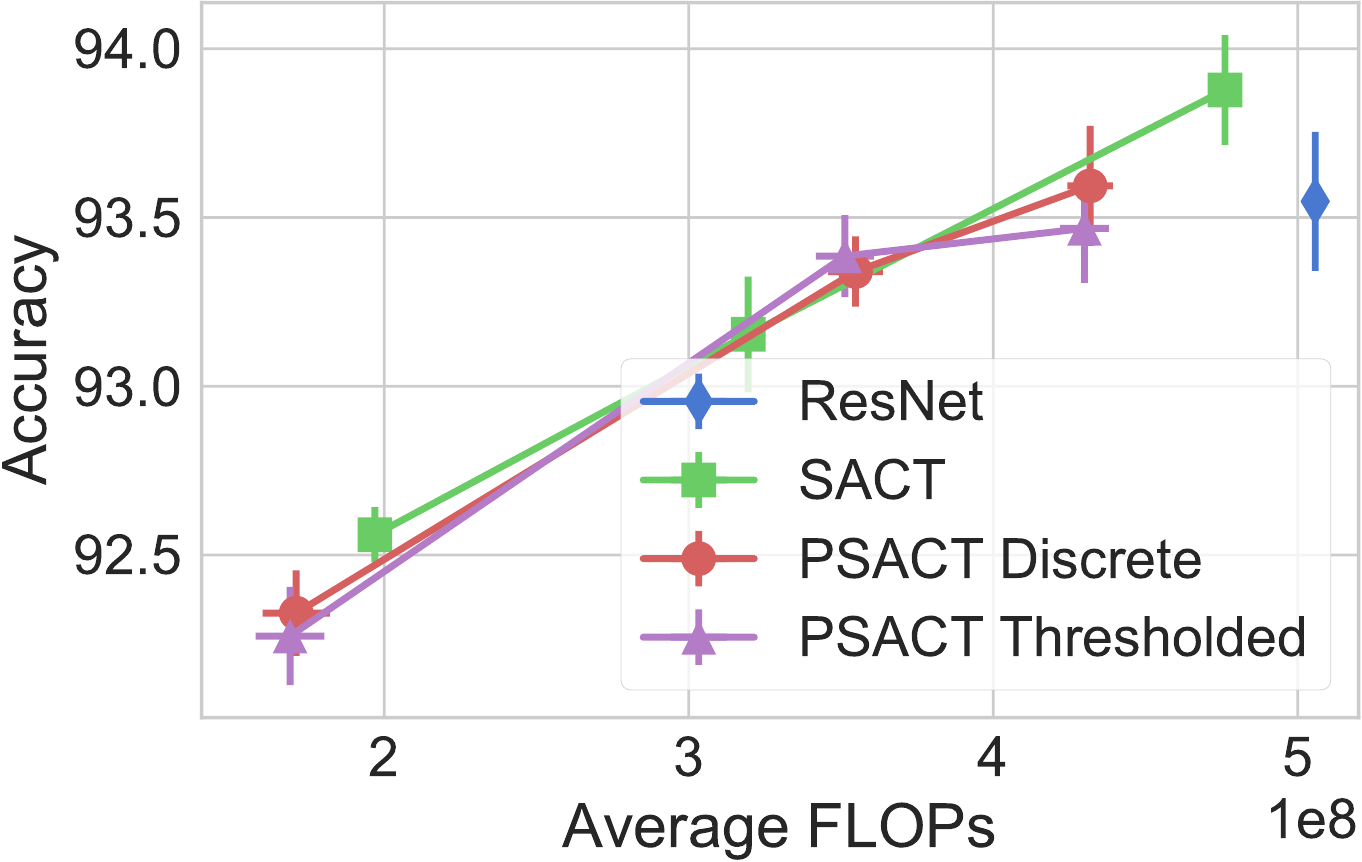}
\end{subfigure}
\caption{Comparison of PSACT (proposed method) and SACT~\cite{figurnov2017spatially} for various values of the computation time penalty $\tau$. PSACT is trained using the relaxed model. The results are averaged over five runs, with error bars denoting one standard deviation. \textbf{Left}: ResNet-32, \textbf{right}: ResNet-110.}
\label{fig:sact-psact}
\end{figure}

\begin{figure}
    \centering
    \footnotesize
    \begin{tabular}{@{\hskip2pt}c@{\hskip2pt}c@{\hskip2pt}c@{\hskip2pt}c@{\hskip2pt}c@{\hskip2pt}c@{\hskip2pt}c}
        \includegraphics[width=0.15\linewidth]{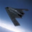} & \includegraphics[width=0.15\linewidth]{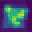} & \includegraphics[width=0.15\linewidth]{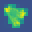} &
        \includegraphics[width=0.15\linewidth]{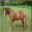} & \includegraphics[width=0.15\linewidth]{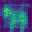} & \includegraphics[width=0.15\linewidth]{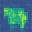} & \includegraphics[width=0.0435\linewidth]{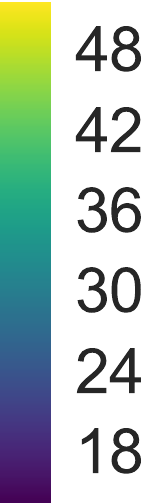} \\
    \end{tabular}
    \caption{Left to right: CIFAR-10 test image, ponder cost map for SACT, expected number of residual units per spatial position for PSACT. SACT and PSACT are applied to ResNet-110 with $\tau=0.005$. Both methods dedicate the computation to the object of interest.}
    \label{fig:ponder_cost}
\end{figure}

Finally, we compare SACT and PSACT models for ResNet-32 and ResNet-110 on fig. \ref{fig:sact-psact}.
The PSACT model is trained using the relaxation and then evaluated in the discrete and thresholded regimes.
PSACT and SACT perform similarly.
We find that PSACT requires using somewhat lower computation time penalty $\tau$ to achieve the same number of FLOPs, perhaps because the expected number of iterations penalty in PSACT is easier to optimize than the surrogate ponder cost of SACT.
Relaxed PSACT successfully trains on ResNet-110, where we have $M=1344$ 18-ary discrete latent variables.
PSACT can be evaluated in deterministic Thresholded mode with very close results, indicating that the latent variables probabilities have saturated.
This is not the case for SACT: evaluation in Thresholded mode reduces the accuracy by at least 5\% (a plot is available in the supplementary materials).
We also present the comparison of the learned computation time maps on fig. \ref{fig:ponder_cost}.

\section{Conclusion}

We have presented Probabilistic Adaptive Computation Time, a principled latent variable model for varying the amount of computation in deep models.
The proposed stochastic variational optimization allows to perform approximate MAP inference in this model.
Experimentally, we find that training using Concrete relaxation of discrete latent variables outperforms REINFORCE-based training.
The model achieves similar results to the heuristic method Adaptive Computation Time, while enjoying a principled formulation.
It can also be used in Thresholded mode with a very simple test-time behavior.
In future, we plan to explore different training techniques and modifications of the proposed latent variable model.
Additionally, we expect that the proposed techniques could be useful for replacing REINFORCE in training of hard attention models.

{
\textbf{Acknowledgments.}
M. Figurnov and D. Vetrov are supported by Russian Science Foundation grant 17-71-20072 and Russian Academic Excellence Project `5-100'.
}

{
\bibliographystyle{plain}
\bibliography{main}
}

\clearpage
\appendix

\begin{strip}
\begin{center}
\textbf{\Large Supplementary materials}
\end{center}
\end{strip}

\section{Algorithms for adaptive computation blocks}

We present the algorithms for discrete adaptive computation block in alg.~\ref{alg:discrete-pact}, for thresholded block in alg.~\ref{alg:thresholded-pact} and for relaxed block in alg.~\ref{alg:relaxed-pact}.
Additionally, the adaptive computation time relaxation for the block is presented in alg.~\ref{alg:act}.
We see that discrete and thresholded blocks allow more straightforward implementation than the adaptive computation time mechanism.

\begin{figure}
\centering
\footnotesize
\begin{tabular}{@{\hskip2pt}c@{\hskip2pt}c@{\hskip2pt}c@{\hskip2pt}c@{\hskip2pt}c@{\hskip2pt}c@{\hskip2pt}c}
\includegraphics[width=0.15\linewidth]{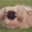} &
\includegraphics[width=0.15\linewidth]{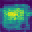} &
\includegraphics[width=0.15\linewidth]{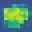} &
\includegraphics[width=0.15\linewidth]{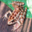} &
\includegraphics[width=0.15\linewidth]{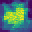} &
\includegraphics[width=0.15\linewidth]{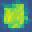} &
\includegraphics[width=0.0435\linewidth]{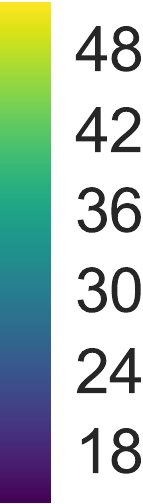} \\
\includegraphics[width=0.15\linewidth]{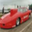} &
\includegraphics[width=0.15\linewidth]{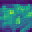} &
\includegraphics[width=0.15\linewidth]{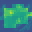} &
\includegraphics[width=0.15\linewidth]{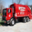} &
\includegraphics[width=0.15\linewidth]{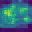} &
\includegraphics[width=0.15\linewidth]{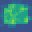} &
\includegraphics[width=0.0435\linewidth]{pics/ponder_maps_extended/colorbar-crop.pdf} \\
\includegraphics[width=0.15\linewidth]{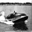} &
\includegraphics[width=0.15\linewidth]{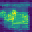} &
\includegraphics[width=0.15\linewidth]{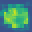} &
\includegraphics[width=0.15\linewidth]{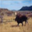} &
\includegraphics[width=0.15\linewidth]{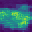} &
\includegraphics[width=0.15\linewidth]{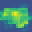} &
\includegraphics[width=0.0435\linewidth]{pics/ponder_maps_extended/colorbar-crop.pdf} \\
\includegraphics[width=0.15\linewidth]{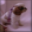} &
\includegraphics[width=0.15\linewidth]{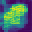} &
\includegraphics[width=0.15\linewidth]{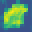} &
\includegraphics[width=0.15\linewidth]{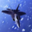} &
\includegraphics[width=0.15\linewidth]{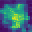} &
\includegraphics[width=0.15\linewidth]{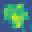} &
\includegraphics[width=0.0435\linewidth]{pics/ponder_maps_extended/colorbar-crop.pdf} \\
\includegraphics[width=0.15\linewidth]{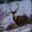} &
\includegraphics[width=0.15\linewidth]{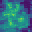} &
\includegraphics[width=0.15\linewidth]{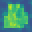} &
\includegraphics[width=0.15\linewidth]{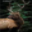} &
\includegraphics[width=0.15\linewidth]{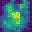} &
\includegraphics[width=0.15\linewidth]{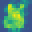} &
\includegraphics[width=0.0435\linewidth]{pics/ponder_maps_extended/colorbar-crop.pdf} \\
\includegraphics[width=0.15\linewidth]{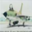} &
\includegraphics[width=0.15\linewidth]{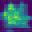} &
\includegraphics[width=0.15\linewidth]{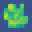} &
\includegraphics[width=0.15\linewidth]{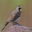} &
\includegraphics[width=0.15\linewidth]{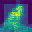} &
\includegraphics[width=0.15\linewidth]{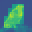} &
\includegraphics[width=0.0435\linewidth]{pics/ponder_maps_extended/colorbar-crop.pdf} \\
\includegraphics[width=0.15\linewidth]{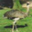} &
\includegraphics[width=0.15\linewidth]{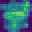} &
\includegraphics[width=0.15\linewidth]{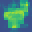} &
\includegraphics[width=0.15\linewidth]{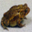} &
\includegraphics[width=0.15\linewidth]{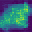} &
\includegraphics[width=0.15\linewidth]{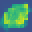} &
\includegraphics[width=0.0435\linewidth]{pics/ponder_maps_extended/colorbar-crop.pdf} \\
\includegraphics[width=0.15\linewidth]{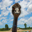} &
\includegraphics[width=0.15\linewidth]{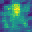} &
\includegraphics[width=0.15\linewidth]{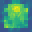} &
\includegraphics[width=0.15\linewidth]{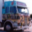} &
\includegraphics[width=0.15\linewidth]{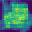} &
\includegraphics[width=0.15\linewidth]{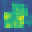} &
\includegraphics[width=0.0435\linewidth]{pics/ponder_maps_extended/colorbar-crop.pdf} \\
\includegraphics[width=0.15\linewidth]{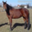} &
\includegraphics[width=0.15\linewidth]{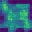} &
\includegraphics[width=0.15\linewidth]{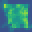} &
\includegraphics[width=0.15\linewidth]{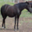} &
\includegraphics[width=0.15\linewidth]{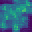} &
\includegraphics[width=0.15\linewidth]{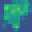} &
\includegraphics[width=0.0435\linewidth]{pics/ponder_maps_extended/colorbar-crop.pdf} \\
\includegraphics[width=0.15\linewidth]{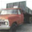} &
\includegraphics[width=0.15\linewidth]{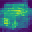} &
\includegraphics[width=0.15\linewidth]{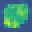} &
\includegraphics[width=0.15\linewidth]{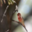} &
\includegraphics[width=0.15\linewidth]{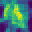} &
\includegraphics[width=0.15\linewidth]{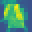} &
\includegraphics[width=0.0435\linewidth]{pics/ponder_maps_extended/colorbar-crop.pdf}
\end{tabular}
    \caption{Left to right: CIFAR-10 test image, ponder cost map for SACT, expected number of residual units per spatial position for PSACT. SACT and PSACT are for ResNet-110 with $\tau=0.005$. The test images were chosen randomly.}
    \label{fig:ponder-cost-extended}
\end{figure}

\begin{figure*}
\centering
\begin{subfigure}[t]{0.4\linewidth}
   \centering
   \includegraphics[width=\linewidth]{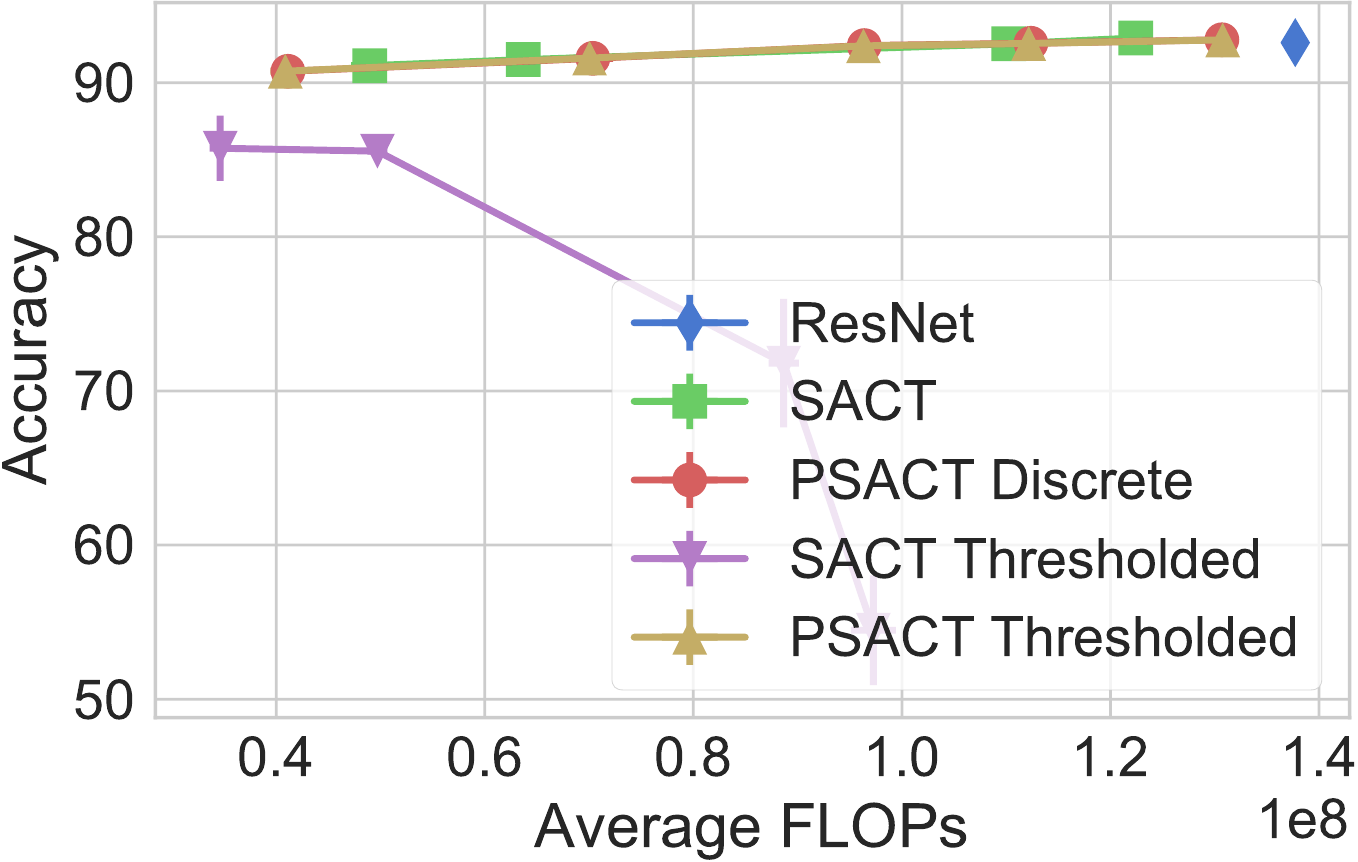}
\end{subfigure}
\quad
\begin{subfigure}[t]{0.4\linewidth}
   \centering
   \includegraphics[width=\linewidth]{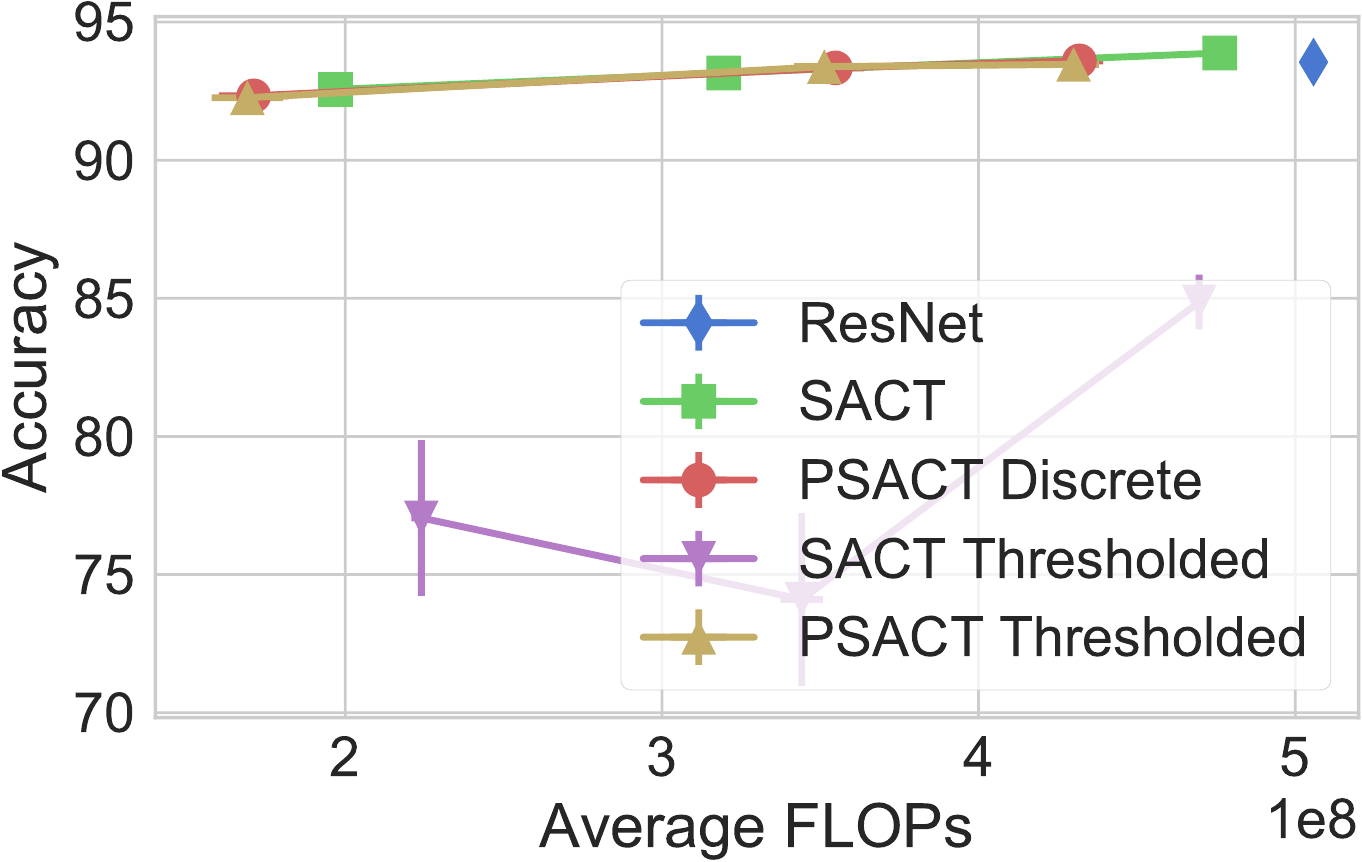}
\end{subfigure}
\caption{Extended version of figure 5. SACT and PSACT for various values of $\tau$. SACT Thresholded is a model trained with SACT and evaluated as a PSACT Thresholded model. \textbf{Left}: ResNet-32, \textbf{right}: ResNet-110.}
\label{fig:sact-psact-extended}
\end{figure*}

\section{Training hyperparameters and additional experimental results}

The training hyperparameters are as follows.
The batch size is 128, weight decay is 0.0002.
The training is performed for 100,000 iterations.
The weights are initialized with variance scaling initializer \cite{he2015delving}.
For all the experiments, except training using REINFORCE, we use SGD optimizer with momentum 0.9.
The initial learning rate is 0.1, decayed by a factor of 10 after 60,000, 75,000 and 90,000 training iterations.
For training of SACT and PSACT models, we use the initialization heuristics from~\cite{figurnov2017spatially} to prevent the \textit{dead residual units} problem.
Namely, we initialize the weights of model with a pretrained vanilla ResNet, and initialize the biases of the logits of the halting probabilities with a constant $-3$.
We train the relaxed PSACT models with temperature $\lambda = \nicefrac{2}{3}$ and clipping threshold $\delta=0.01$.
We have explored temperatures in the range $\lambda \in [0.4, 0.8]$ and obtain similar results.

We demonstrate additional examples of the computation time maps of SACT and PSACT in fig.~\ref{fig:ponder-cost-extended}.

An extended version of figure 5 from the main text is shown on fig.~\ref{fig:sact-psact-extended}.
We demonstrate that when a model trained with SACT relaxation is evaluated as a PSACT Thresholded model, the accuracy significantly drops.
This indicates that training using SACT does not result in a sharp halting distribution.

The values of $\tau$ in this experiment are as follows.
ResNet-32 PSACT: $\tau \in \{5e-2, 2e-2, 1e-2, 5e-3, 1e-3 \}$
ResNet-32 SACT: $\tau \in \{1e-1, 5e-2, 1e-2, 5e-3\}$.
ResNet-110  PSACT: $\tau \in \{1e-2, 5e-3, 1e-3\}$.
ResNet-110 SACT: $\tau \in \{5e-3, 1e-3, 5e-4\}$.
Higher values of $\tau$ correspond to less FLOPs.

\begin{algorithm}[t]
\begin{algorithmic}[1]
    \Require maximum number of iterations $L$
    \Ensure number of executed iterations $z$
    \Ensure output of the block
    \For{$l = 1 \dots L$}
        \State Compute $u^l$
        \If{$l < L$} $h = H^l (u^l)$
        \Else $\ h = 1$
        \EndIf
        \State $\xi \sim \operatorname{Bernoulli}(h)$
        \If{$\xi = \mathbf{true}$}
            \State $\mathrm{output} = u^l$
            \State $z = l$
            \State \Return $\mathrm{output},\ z$
        \EndIf
    \EndFor
\end{algorithmic}
\caption{Discrete adaptive computation block.}
\label{alg:discrete-pact}
\end{algorithm}

\begin{algorithm}[t]
\begin{algorithmic}[1]
    \Require maximum number of iterations $L$
    \Ensure number of executed iterations $z$
    \Ensure output of the block
    \For{$l = 1 \dots L$}
        \State Compute $u^l$
        \If{$l < L$} $h = H^l (u^l)$
        \Else $\ h = 1$
        \EndIf
        \If{$h > 0.5$}
            \State $\mathrm{output} = u^l$
            \State $z = l$
            \State \Return $\mathrm{output},\ z$
        \EndIf
    \EndFor
\end{algorithmic}
\caption{Thresholded adaptive computation block.}
\label{alg:thresholded-pact}
\end{algorithm}

\begin{algorithm}[t]
\begin{algorithmic}[1]
    \Require maximum number of iterations $L$
    \Require temperature of relaxation $\lambda$
    \Ensure expected number of iterations $N$
    \Ensure output of the block
    \State $S_{\hat{\xi}} = 1$ \Comment{Remaining stick length for $\hat{\xi}$}
    \State $S_h = 1$ \Comment{Remaining stick length for $h$}
    \State $N = 0$
    \State $\widehat{\mathrm{output}} = 0$
    \For{$l = 1 \dots L$}
        \State Compute $u^l$
        \If{$l < L$} $h = H^l (u^l)$
        \Else $\ h = 1$
        \EndIf
        \State $\hat{\xi} \sim \operatorname{RelaxedBernoulli}(h; \lambda)$
        \State $\hat{z} = S_{\hat{\xi}} \cdot \hat{\xi}$
        \State $\widehat{\mathrm{output}} = \widehat{\mathrm{output}} + \hat{z} \cdot u^l$
        \State $N = N + l \cdot S_h \cdot h$
        \State $S_{\hat{\xi}} = S_{\hat{\xi}} (1 - \hat{\xi})$
        \State $S_h = S_h (1 - h)$
    \EndFor
    \State \Return $\mathrm{output},\ N$
\end{algorithmic}
\caption{Relaxed adaptive computation block.}
\label{alg:relaxed-pact}
\end{algorithm}

\begin{algorithm}[t]
\begin{algorithmic}[1]
    \Require maximum number of iterations $L$
    \Require $0 < \varepsilon < 1$ \Comment{Recommended value: $0.01$}
    \Ensure ponder cost $\rho$ \Comment{Upper bound on the number of executed iterations}
    \Ensure output of the block
    \State $c = 0$ \Comment{Cumulative halting probability}
    \State $R = 1$ \Comment{Remainder}
    \State $\mathrm{output} = 0$
    \State $\rho = 0$
    \For{$l = 1 \dots L$}
        \State Compute $u^l$
        \If{$l < L$} $h = H^l (u^l)$
        \Else $\ h = 1$
        \EndIf
        \State{$c = c + h$}
        \State{$\rho = \rho + 1$}
        \If{$c < 1-\varepsilon$}
            \State{$\mathrm{output} = \mathrm{output} + h \cdot u^l$}
            \State{$R = R - h$}
        \Else
            \State{$\mathrm{output} = \mathrm{output} + R \cdot u^l$}
            \State{$\rho = \rho + R$}
            \State \textbf{break}
        \EndIf
    \EndFor
    \State \Return $\mathrm{output}, \rho$
\end{algorithmic}
\caption{Adaptive computation block with Adaptive Computation Time relaxation.}
\label{alg:act}
\end{algorithm}

\end{document}